\documentclass[letterpaper, 10pt, conference]{ieeeconf}

\IEEEoverridecommandlockouts
\let\labelindent\relax

\usepackage[utf8]{inputenc}
\usepackage{amsfonts}
\usepackage[inline]{enumitem}
\usepackage{soul}
\usepackage{color}
\usepackage{multirow}
\usepackage{cite}
\usepackage{graphicx}
\graphicspath{{./figures/}}
\usepackage{amsmath}
\usepackage{amssymb}
\usepackage{pifont}
\newcommand{\cmark}{\ding{51}} 
\newcommand{\xmark}{\ding{55}} 
\usepackage{bm}
\usepackage[font=small,labelfont=bf,justification=justified]{caption}
\usepackage{subcaption}
\usepackage{multicol}
\usepackage{flushend}
\usepackage[bookmarks=true]{hyperref}
\usepackage[ruled,vlined,linesnumbered]{algorithm2e}
\usepackage{booktabs}   
\usepackage{graphicx}

\renewcommand{\baselinestretch}{0.986}

\RestyleAlgo{ruled}
\title{Manipulation Planning for Construction Activities with Repetitive Tasks}

\author{}
\date{}

\author{Wangyi Liu$^1$, Dasharadhan Mahalingam$^1$, Fanru Gao$^2$, Ci-Jyun Liang$^2$, and Nilanjan Chakraborty$^1$
\thanks{$^{1}$Department of Mechanical Engineering, Stony Brook University, NY, USA.
{\tt\small \{wangyi.liu, dasharadhan.mahalingam, nilanjan.chakraborty\}@stonybrook.edu.}}
\thanks{$^{2}$Department of Civil Engineering, Stony Brook University, NY, USA.
{\tt\small \{fanru.gao, ci-jyun.liang\}@stonybrook.edu.}}%
}

\begin{document}

\maketitle

\begin{abstract}
In this paper, we study the problem of manipulation skill acquisition for performing construction activities consisting of repetitive tasks (e.g., building a wall or installing ceiling tiles). Our approach involves setting up a simulated construction activity in a Virtual Reality (VR) environment, where the user can provide demonstrations of the object manipulation skills needed to perform the construction activity. We then exploit the screw geometry of motion to approximate the demonstrated motion as a sequence of constant screw motions. For performing the construction activity, we generate the sequence of manipulation task instances and then compute the joint space motion plan corresponding to each instance using Screw Linear Interpolation (ScLERP) and Resolved Motion Rate Control (RMRC). We evaluate our framework by executing two representative construction tasks: constructing brick walls and installing multiple ceiling tiles. Each task is performed using only a single demonstration, a pick-and-place action for the bricks, and a single ceiling tile installation. Our experiments with a 7-DoF robot in both simulation and hardware demonstrate that the approach generalizes robustly to arbitrarily long construction activities that involve repetitive motions and demand precision, even when provided with just one demonstration. For instance, we can construct walls of arbitrary layout and length by leveraging a single demonstration of placing one brick on top of another.

\textit{Video}--- \url{https://youtu.be/SEwtaH--NZg}
\end{abstract}

\section{Introduction}
In the construction domain, there is a strong need for robotic systems capable of executing repetitive tasks characterized by precision, stringent geometric constraints, and extended temporal horizons. Examples include bricklaying, ceiling tile installation, beam alignment, and component assembly. These tasks are inherently challenging because they impose strict kinematic constraints on the robot’s end-effector trajectory and demand repeatable, high-accuracy execution over long sequences of actions. Consider the case of bricklaying: each brick must be positioned with its bottom face parallel to the ground and precisely aligned with respect to the previously placed bricks, ensuring structural integrity and layout conformity. Similarly, ceiling tile installation requires continuous adjustment of the tile’s pose, i.e., position and orientation, relative to the ceiling frame during insertion. 

These activities can be abstracted as sequences of repeated, parameterized motion primitives; a single pick-and-place operation for a brick, or the insertion of a single ceiling tile. Consequently, if a robotic system can successfully parameterize and execute one instance of such a primitive, and is provided with a specification of the wall layout or ceiling grid, it should, in principle, be capable of autonomously generating and executing the full sequence required to complete the entire construction task.
Furthermore, the spatial extent of the task, e.g., wall length, grid size, or layout topology, may be arbitrarily large. In this work, we \emph{formalize and investigate the problem of generalizing from a single demonstration of a base task to the execution of repetitive construction activities of arbitrary length, using bricklaying and ceiling tile installation as representative exemplars}.

\begin{figure}[!t]
    \centering
    \includegraphics[width=\linewidth]{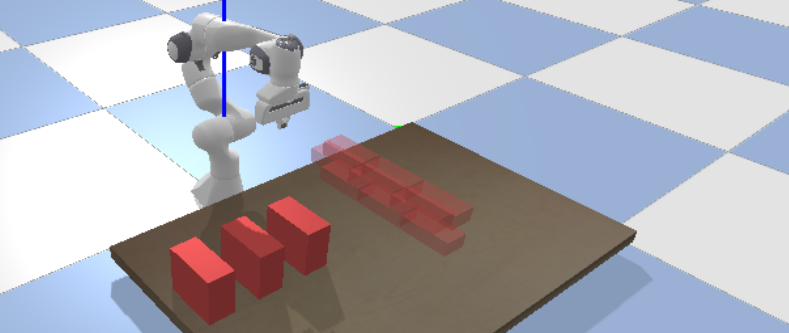}
    \caption{
    Construction of a three-layer wall with a total of nine bricks in a simulation environment. The bricks are stacked in a pile initially (solid red bricks on the left). The generated task instances show the goal poses of the bricks for constructing a wall, visualized as translucent bricks on the right.}
    \label{fig:construction_activity_setup}
\end{figure}

We formulate the construction problem within the paradigm of programming by demonstration (PbD)~\cite{billard2016learning}, assuming that task demonstrations are provided in a Virtual Reality (VR) environment. Multiple modalities exist for acquiring demonstrations for robotic learning, including teleoperation, kinesthetic teaching, VR-based demonstrations, demonstrations by skilled human workers, and even video-based demonstrations. Teleoperation typically requires specialized hardware and operator training, while kinesthetic demonstrations are difficult to apply in construction contexts due to the complexity of setting up realistic physical scenarios with heavy or fixed building materials. Human worker demonstrations are expensive and time-consuming to collect at scale, as they require training personnel to act as reliable data sources. Video demonstrations, although easy to capture, do not provide object trajectories directly in $SE(3)$, the Lie group of rigid body transformations, which are essential for precise reproduction. In contrast, VR-based demonstrations present a compelling alternative: they are cost-effective to collect, require only a VR headset and controllers, and directly yield object trajectories as continuous curves in $SE(3)$. These trajectories naturally encode the geometric and kinematic constraints that govern the task, which we model as one-parameter subgroups of $SE(3)$, or equivalently, as constant-pitch screw motions~\cite{mahalingam2023humanguided}. This representation captures the essential invariants (or constraints) of the motion and enables generalization across task instances.

{\bf Contributions}: We present a framework that enables a generalization of arbitrary-length construction tasks from a single step-level demonstration. Rather than requiring a full demonstration of the entire wall-building or ceiling-tiling process, our approach only needs a single instance of the elemental task, e.g., placing one brick or installing one tile, captured in a VR environment. The object trajectory in 
$SE(3)$ is then segmented into constant screw motions, and the underlying kinematic constraints are extracted and parameterized. For new task instances with arbitrary layouts or lengths, these parameterized screw motions are instantiated relative to the new object poses, and motion plans are generated using ScLERP~\cite{ScLERP} and Resolved Motion Rate Control (RMRC)~\cite{Whitney1969ResolvedMotion}, which ensures that we obtain a joint space path that preserves the demonstrated task space constraints. This enables the robot to autonomously replicate the demonstrated step across an arbitrary number of repetitions, effectively constructing entire walls or completing full ceiling installations. Our approach thus provides a highly data-efficient solution for scaling single-step demonstrations to arbitrarily long and spatially varied construction tasks.

\section{Related Work}
Robotics has been increasingly adopted in the construction domain to automate labor-intensive and repetitive tasks. Early work by \cite{MasonryPrototype} introduced one of the first masonry robot prototypes, focusing on the design of human–robot collaborative masonry systems. In the construction industry, Construction Robotics developed the SAM system \cite{SAM}, which automates large-scale, on-site bricklaying and significantly reduces manual workload while increasing productivity. However, such systems are typically designed for a single type of construction task (e.g., wall building).

To enable automation across diverse construction 
tasks, researchers have explored learning-based methods such as Learning from Demonstration (LfD) \cite{LfDSurvey} and reinforcement learning (RL) \cite{ReinforLearning}. LfD uses human-provided demonstrations to train robots to perform new skills. For instance, \cite{LiangLfD,BeamAssemble} have employed LfD to teach robots to install ceiling tiles in pre-assembled grids and to perform steel beam assembly. RL approaches, on the other hand, learn task policies by optimizing a reward function and have been applied to multi-robot construction \cite{multiRobotConstrct}, earth-moving \cite{earthMoving}, and pick-and-place tasks \cite{RL-VR-pickandplace}. While promising, both LfD and RL approaches generally require a large number of demonstrations or rollouts to achieve robust performance, which limits their scalability in real-world construction scenarios \cite{chandramouliconstruction}.  

Motion Primitive based approaches \cite{gutierrez2025movement} have also been used to learn tasks through demonstrations. While GMM\cite{calinon2014tpgmm} based approaches require multiple demonstrations to learn tasks, DMP\cite{IjspeertNHPS13} based approaches learn from a single task. A major drawback for such approaches is that it is not clear if a single choice of hyper-parameters can be transferred across a wide variety of tasks. Moreover tuning these parameters becomes even harder when the end user is not a subject expert. Screw Primitive based approaches such as \cite{screwmimic} have been proposed for learning tasks. But they restrict themselves to tasks with only pure rotation or pure translation motion.

\emph{This motivates the development of approaches that can generalize from a single step-level demonstration by learning a parameterized skill representation that can be reused across arbitrarily many task repetitions, which is the central focus of our work.}

A critical aspect of LfD-based construction automation is the method of demonstration acquisition. Existing forms of demonstrations include video demonstrations, force-based demonstrations, and trajectory demonstrations \cite{LiangLfD}. These can be collected using two primary approaches: (i) \emph{physical guidance}, in which a human manually moves the robot (or its end-effector) to perform the task while its joint trajectories are recorded, and (ii) \emph{visual guidance}, which uses motion capture systems or cameras to record a human expert’s motion and infer object or end-effector trajectories \cite{LfDRobotSurvey}.  

Recent advances in VR have provided a compelling alternative for demonstration acquisition \cite{VRdemon1,VRdemon2,VRdemon3}. Compared to traditional approaches, VR offers a safe, low-cost, and repeatable means of capturing demonstrations \cite{VRRobot}.
VR allows a first-person view of task execution and enables real-time interaction with objects in a simulated environment \cite{VRNavigate}.
When used for robot learning, VR has been shown to reduce the number of policy-learning steps \cite{kawakami2021VRlearning} and to generate high-quality demonstrations for training deep neural networks \cite{jackson2019VRbenefits}. Recent work has applied VR to teach robots to perform tasks such as window installation \cite{huang2022VRWindow} and to analyze collaborative human–robot construction scenarios involving drones and robotic arms \cite{NunoVRanalysis}. VR based demonstration acquisition is thus emerging as an important tool for advancing autonomous construction \cite{huang2022rlconstruction,li2023construction,wang2021digitaltwin}.  

The approach proposed in \cite{mahalingam2023humanguided} uses kinesthetic demonstrations to solve the problem of motion generation for complex manipulation tasks. Building on this idea, we demonstrate that activities composed of repetitive, constraint-rich manipulation tasks can be parameterized and executed robustly using demonstrations collected in a VR environment.

\emph{To the best of our knowledge, this is the first work that combines VR-based demonstration with screw-theoretic motion representation to enable step-level generalization and scalable execution of arbitrarily long construction activities.}

\begin{figure*}[htbp]
    \centering
    \includegraphics[width=0.75\textwidth]{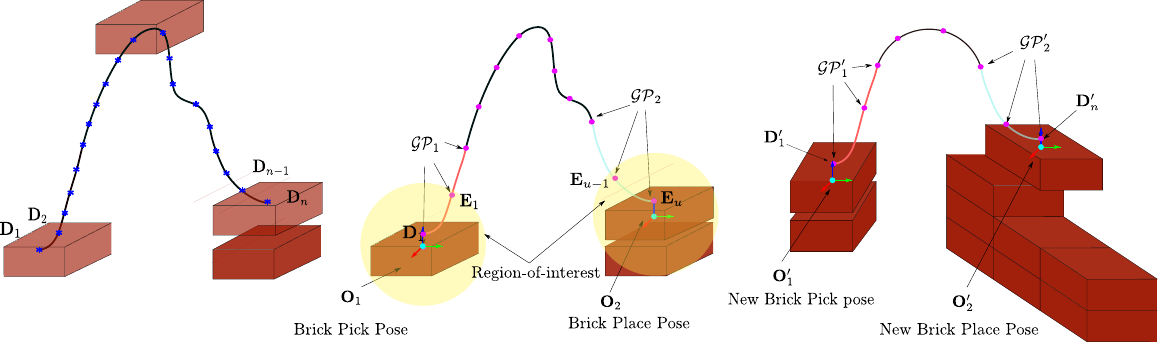}
    \caption{\textbf{\textsc{Schematic Sketch of Motion Estimation for Building a Wall:}} \textbf{Left} - The collected demonstrations $\mathbf{D}$ in VR which consists of a sequence of $SE(3)$ poses which are represented with blue markers. \textbf{Center} -  Segmenting the demonstration $\mathbf{D}$ as a sequence of constant screw motions, $\{\mathbf{D}_1, \mathbf{E}_1, \dots, \mathbf{E}_u\}$, the \emph{``Key Segments"} ($\mathcal{GP}_1$ and $\mathcal{GP}_2$) are determined based on the region-of-interest centered at the initial and final object poses, $\mathbf{O}_1$ and $ \mathbf{O}_2$ respectively. \textbf{Right} - Given new initial and final object poses, $\mathbf{O}_1'$ and $ \mathbf{O}_2'$, the \emph{``Guiding Poses"} $\mathcal{GP}_1'$ and $\mathcal{GP}_2'$ are determined by transforming $\mathcal{GP}_1$ and $\mathcal{GP}_2$ with respect to the new object poses.} 
    \label{fig:schematic_sketch}
    \vspace{2mm}
\end{figure*}

\section{Mathematical Preliminaries}
In this section, we present a brief review of the mathematical background required to understand this work.

\noindent
\textbf{Screw Displacement:}
The Chasles-Mozzi theorem states that the general Euclidean displacement/motion of a rigid body from the origin $\mathbf{I}$ to $\mathbf{T} = (\mathbf{R},\boldsymbol{p}) \in SE(3)$
can be expressed as a rotation $\theta$ about a fixed axis $\mathcal{S}$, called the \textit{screw axis}, and a translation $d$ along that axis. 
Plücker coordinates can be used to represent the screw axis by $\boldsymbol{\omega}$ and $\boldsymbol{m}$, where $\boldsymbol{\omega} \in \mathbb{R}^3$ is a unit vector that represents the direction of the screw axis, $\boldsymbol{m} = \boldsymbol{r} \times \boldsymbol{\omega}$, and $\boldsymbol{r} \in \mathbb{R}^3$ is an arbitrary point on the screw axis. Thus, the screw parameters are defined as $\boldsymbol{\omega}, \boldsymbol{m}, h, \theta$, where $h$ is the pitch of the screw and $\theta$ is its magnitude. In general, for pure rotation and general screw motion, $h$ is finite, while for pure translation, $h = \infty$.
If $\mathbf{R} \neq \mathbf{I}$, then by using the standard procedure to obtain the rotation axis and magnitude from the rotation matrix $\mathbf{R}$, we can determine $\bm{\omega}$ and $\theta$. The pitch is given by $h = \bm{\omega}^T\bm{\upsilon}$ and $\mathbf{m} = \bm{\upsilon} - h\bm{\omega}$, where $\bm{\upsilon} = \left[(\mathbf{I} - e^{\hat{\bm{\omega}}\theta})\hat{\bm{\omega}} + \theta\bm{\omega}\bm{\omega}^T\right]^{-1}\bm{p}$. If $\mathbf{R} = \mathbf{I}$, then the motion is pure translation, where $h = \infty$ and $\bm{m} = \bm{0}$ by definition. We can obtain $\theta$ and $\bm\omega$ from $\theta = ||\bm{p}||$ and $\bm{\omega} = \bm{p}/||\bm{p}||$.
\textbf{A {\em constant screw motion} is a motion where the parameters $\boldsymbol{\omega}, \boldsymbol{m}$, and $h$ stay constant throughout the motion}.

Given the screw parameters $\boldsymbol{\omega}, \boldsymbol{m}, h$, the screw displacement for a motion of magnitude $\theta$ can be obtained using the matrix exponential, $\mathbf{T} = e^{\widehat{\bm\xi}\theta}$. Here, $\widehat{\bm\xi} \in se(3)$ and $\bm\xi \in \mathbb{R}^6$ are the unit twist and unit twist coordinates associated with the motion. They are defined as,
\begin{gather}
    \label{eq:unit_twist_general_screw}
    \widehat{\bm{\xi}} =
    \begin{bmatrix}
        \widehat{\bm{\omega}} & \bm{m} + h\bm{\omega} \\ 0 & 0
    \end{bmatrix}, \bm{\xi} = 
    \begin{bmatrix}
        \bm{m} + h\bm{\omega} \\ \bm{\omega}
    \end{bmatrix}
    \text{for}~h\neq\infty \\
    \label{eq:unit_twist_prismatic}
    \widehat{\bm{\xi}} =
    \begin{bmatrix}
        \mathbf{I} & \bm{\omega} \\ 0 & 0
    \end{bmatrix}, \bm{\xi} = 
    \begin{bmatrix}
        \bm{\omega} \\ \bm{0}
    \end{bmatrix}
    \text{for}~h=\infty
\end{gather}
While the matrix exponential maps elements of $se(3)$ to $SE(3)$, the matrix logarithm maps elements of $SE(3)$ to $se(3)$. Please refer to\cite{lynch2017modern} for more details.
\begin{gather}
    e^{\widehat{\bm\xi}\theta} = \mathbf{T},
    \quad
    \log{\mathbf{T}} = \widehat{\bm\xi}\theta
\end{gather}

\noindent
\textbf{Screw Linear Interpolation (ScLERP):}
To perform a one degree-of-freedom smooth screw motion
between two object poses in $SE(3)$, ScLERP can be used. ScLERP generates a geodesic motion between two given poses $\mathbf{G}_i, \mathbf{G}_f \in SE(3)$. ScLERP can be performed using,
\begin{gather}
    \label{eq:sclerp}
    \mathbf{G}_\tau = e^{\widehat{\bm\xi}\tau\theta} \mathbf{G}_i
\end{gather}
where, $\widehat{\bm{\xi}}\theta = \log(\mathbf{G}_f {\mathbf{G}^{-1}_i})$ is the twist that results in the screw displacement form $\mathbf{G}_i$ to $\mathbf{G}_f$ and $\tau\in[0,1]$ is the linear interpolation factor.

\noindent
\textbf{Task Instance:} Objects affecting the generation of motion plans for manipulation tasks are defined as task-related objects and the set $\mathcal{O} = \{\mathbf{O}_1, \mathbf{O}_2, \dots, \mathbf{O}_u\}$ containing the $SE(3)$ poses of all the task-related objects is defined as a task instance.

\noindent
\textbf{Demonstration:} A demonstration of a manipulation task is a sequence of $SE(3)$ poses $\mathcal{D} = \langle\mathbf{D}_1, \mathbf{D}_2, \dots, \mathbf{D}_v\rangle$ that defines the motion of the manipulated object.

\noindent
\textbf{SEW Angle:} 
The Shoulder-Elbow-Wrist (SEW) angle $\psi \in \mathbb{R}$, parametrizes the null space of redundant manipulators \cite{seraji1989redundant}. For a common 7-DoF manipulator, the elbow posture is not fully determined by the end-effector pose. It has one extra degree of freedom, making it kinematically redundant which can be exploited to avoid obstacles or joint limits.

\section{Problem Statement}

Consider that we are given a single demonstration $\mathcal{D}$ of an object manipulation skill that is required to perform a repetitive task in a construction activity along with the corresponding task instance $\mathcal{O}$. The problem that we are trying to solve can be stated as: \textbf{Given a demonstration $\mathcal{D}$ of a manipulation skill and its corresponding task instance $\mathcal{O}$, determine the sequence of task instances $\langle \mathcal{O}_1, \mathcal{O}_2, \dots, \mathcal{O}_k \rangle$ and the joint space motion plan $\mathcal{M} = \langle \Theta_1, \Theta_2, \dots, \Theta_l \rangle$ required to transfer the manipulation skill to the new task instances for successfully completing an activity}. For the  construction activity in our work, $\mathcal{D}$ defines the motion that the object goes through during a single task instance of the entire activity, and $\mathcal{O} = \{\mathbf{O}_1, \mathbf{O}_2\}$ consists of the initial pose $\mathbf{O}_1$ and the final pose $\mathbf{O}_2$ of the objects. Coincidentally, $\mathbf{O}_1 = \mathbf{D}_1$ and $\mathbf{O}_2 = \mathbf{D}_v$.

The above problem can be solved by splitting it up into the following three sub-problems:

\noindent
\textbf{Constraint Extraction}:
Given a demonstration $\mathcal{D}$, we need to extract the essential task constraints, i.e., the constraints that characterize the task and should be satisfied for successful execution. 

\noindent
\textbf{Determination of Task Instances}:
For a arbitrary long construction activity, we also need to determine the sequence of task instances which specify the order and pose at which the objects are placed. Additionally, we might need other parameters that define the specification of what to construct. Take the wall building task like an example, we assume that we are given a pose $\mathbf{B} \in SE(3)$ denoting the starting pose of the wall and the specification of the wall in terms of straight, curved or corner along with the number of bricks in each layer ($\beta$), number of layers ($\alpha$), and the horizontal offset between adjacent layers ($\delta$). Since the length$(\ell)$, breadth $(b)$, and width $(h)$ of the brick are known, using this information, we can determine the sequence of task instances that need to be executed to complete the entire construction activity.

\noindent
\textbf{Motion Estimation}:
After computing the sequence of task instances, we now have to determine the joint space motion plan that can successfully perform the entire construction activity while satisfying the extracted motion constraints.

\section{Solution Approach}
\noindent
\textbf{Demonstration Acquisition:} We collect demonstrations of manipulation tasks in a VR environment. The VR environment was built in Unreal Engine 4 (UE4) \cite{ue4} and allows the user to interact with objects in the VR environment by means of a VR headset. We used the Meta Quest 2 \cite{meta_quest} headset in our work. 

\begin{figure*}[!t]
    \centering
    \includegraphics[width=\textwidth]{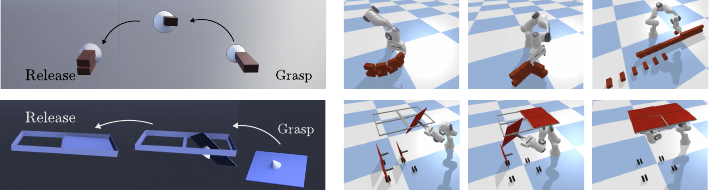}
    \caption{\textbf{\textsc{Demonstration Acquisition and Construction Task Execution:}} The first and second row correspond to wall construction and ceiling tile installation experiments respectively. The left part shows the process of demonstration collection in VR, the right part shows the corresponding experiments in simulation. The following experiments are shown: curved wall construction, corner wall construction, long wall construction and ceiling tile installation.}
    \label{fig:simu_experiments}
\end{figure*}

Two environments are set up (See Figure \ref{fig:simu_experiments}): the first one contains two bricks, one that the user can pick up and manipulate and the other is placed at another location denoting the start of the wall. The user can provide a demonstration by picking up the movable brick and placing it next to the first brick. The second environment contains a flat cuboid to represent the ceiling tile, and a square frame denote the ceiling frame where the ceiling should be installed to. The user provides a demonstration by grasping the tile using the handheld controllers, reorients it, and seats the tile within the frame. The motion of the brick and tile that are being manipulated is recorded and used as the demonstration. This demonstration implicitly captures the constraints required to finish a single task in a construction activity without collision.

\noindent
\textbf{Extraction of Task Constraints:} From the provided demonstration, we then extract the task constraints as a sequence of constant screw motion constraints. Following the approach proposed in \cite{mahalingam2023humanguided}, we are able to extract the motion invariants and determine the object motion for a new task instance.
The recorded demonstration $\mathcal{D}$ is segmented as a sequence of constant screw segments and the constraints associated with the task are extracted as the sequence of guiding poses $\langle\mathbf{G}_1, \mathbf{G}_2, \dots\rangle$ as shown in Figure \ref{fig:schematic_sketch}. For more details please refer to \cite{mahalingam2023humanguided}.
These constraints are independent of the choice of coordinate frame and allow us to successfully transfer the extracted constraints to a new task instance.

\noindent
\textbf{Generation of Task Instances:} In this work, the entire construction activity consists of multiple repetitive tasks. We generate the task instances to perform the construction activity by following deterministic patterns such as lines, curves or polylines once the location of the construction activity is known. 

Consider that the dimension of the construction objects (brick/ceiling tile) along its length, breadth and width are $\ell, b, \text{and } h$ respectively. Let $\epsilon_\ell$, $\epsilon_b$ and $\epsilon_h$ be the spacing required between adjacent objects in the same layer along $x$ and $y$ direction and adjacent layers respectively. Given the starting pose of the construction activity, $\mathbf{B}$, depending on the deterministic format of construction layout, we can also determine the goal poses of different task instances as,
\begin{equation}
\label{eq:taskInstanceDefine}
\begin{cases}
\mathbf{T}_{1,1,1} = \mathbf{B}, \quad  i = 1, j = 1, k = 1\\
\begin{aligned}
\mathbf{T}_{1,1,k}
&= \mathbf{T}_{1,1,k-1}\,\mathbf{Z}(h+\epsilon_h)\,\mathbf{X}\!\big(\Delta_x(i,\delta_x)\big)\\
&\quad \mathbf{Y}\!\big(\Delta_y(i,\delta_y)\big)\,\mathbf{R}(\theta), k>1
\end{aligned}\\
\\
\mathbf{T}_{i,1,k} = \mathbf{T}_{i-1,1,k}\mathbf{Y}(b + \epsilon_b)\mathbf{R}(\theta), i > 1
\\
\mathbf{T}_{1,j,k} = \mathbf{T}_{1,j-1,k}\mathbf{X}(\ell + \epsilon_l)\mathbf{R}(\theta), j > 1

\end{cases}
\end{equation}

Where $\mathbf{X}(t), \mathbf{Y}(t)\text{ and } \mathbf{Z}(t)$ are $SE(3)$ transformations that represent translation along the vectors $\mathbf{\begin{bmatrix}1 & 0 & 0\end{bmatrix}^T}$, $\mathbf{\begin{bmatrix}0 & 1 & 0\end{bmatrix}^T}$ and $\mathbf{\begin{bmatrix}0 & 0 & 1\end{bmatrix}^T}$ respectively by a magnitude of $t$ meters. Here, $i, j$ denote the index of the object along $x, y$ direction respectively and $k$ denotes which layer the object is. Most of the construction tasks are performed along the plane parallel to the ground. Moreover, we assume the construction objects have uniform shape, with top and bottom faces parallel to ground as well. Thus, the orientation $\mathbf{R}(\theta)$ is a $SE(3)$ transformation that represents pure rotation about the axis $\mathbf{\begin{bmatrix}0 & 0 & 1\end{bmatrix}^T}$ by the magnitude $\theta$.
\begin{equation}
\mathbf{X}(t)=\big(\mathbf{I},\,\left[\begin{smallmatrix} t\\0\\0 \end{smallmatrix}\right]\big),\quad
\mathbf{Y}(t)=\big(\mathbf{I},\,\left[\begin{smallmatrix} 0\\t\\0 \end{smallmatrix}\right]\big),\quad
\mathbf{Z}(t)=\big(\mathbf{I},\,\left[\begin{smallmatrix} 0\\0\\t \end{smallmatrix}\right]\big)
\end{equation}
\begin{equation}
    \mathbf{R}(\theta) = \begin{bmatrix}
        \cos{\theta} & -\sin{\theta} & 0 & 0\\
        \sin{\theta} & \cos{\theta} & 0 & 0\\
        0 & 0 & 1 & 0\\
        0 & 0 & 0 & 1
    \end{bmatrix}
\end{equation}
Here, $\Delta(x,y)$ is a real valued function which determines the offset between objects in adjacent layers and for most cases, is defined as,
\begin{equation}
    \Delta(x,y) = 
        \begin{cases}
            y, &\text{if $x$ is even} \\
            0, &\text{otherwise}
        \end{cases}
\end{equation}
\noindent
\textbf{Task Instances for Different Construction Activities:}
For wall construction task, it is generally assumed that it is only one layer thick in the $y$ direction. So, the task instances generated for wall-building type of work is restricted to $x~(horizontal)$ and $z~(vertical)$ directions, in our definition, $j = 1$.

The curvature of the wall is determined by $\theta$. For a straight wall, $\theta = 0$. Setting $\delta = 0$ results in the layers being aligned. 
By modifying equation \eqref{eq:taskInstanceDefine} for the case $i \geq 1, k > 2$ we can also determine the goal pose of the objects for a corner wall,
\begin{equation}
    \label{eq:brick_goal_seq_corner}
    \mathbf{T}_{i, k} =
            \mathbf{T}_{i-1, k}\mathbf{X}(\Delta(i+1, \ell' + \epsilon_\ell))\mathbf{Y}(\Delta(i, \ell' + \epsilon_\ell))\mathbf{R}(\theta)
\end{equation}
with $\theta = 90^{\circ}$ at the corner and $\theta = 0^{\circ}$ everywhere else. Here $\ell' = (\ell+b)/2$. So the wall extends along the $x$ direction before the corner, and continues along the $y$ direction after the corner.

For the ceiling tile installation task, we assume that there is only one layer, restricting this task to only $x$ and $y$ directions. The poses of all ceiling tiles are given by,
\begin{equation}
    \label{eq:ceiling_tile_sqc}
\begin{cases}
\mathbf{T}_{i,1} = \mathbf{T}_{i-1,1}\mathbf{Y}(b + \epsilon_b) &\text{along $x$ direction}\\
\mathbf{T}_{1,j} = \mathbf{T}_{1,j-1}\mathbf{X}(\ell + \epsilon_l) &\text{along $y$ direction}
\end{cases}
\end{equation}

Equations \eqref{eq:taskInstanceDefine} - \eqref{eq:ceiling_tile_sqc} determine the goal pose of the objects given the indices $i$, $j$ and $k$.

Using $\mathbf{T}_{i, j, k}$, we can now determine the sequence of the task instances. Each task instance $\mathcal{O}_k$ consists of the initial pose and the final pose of the brick. Since we assume that the bricks are stacked initial at a known location and the goal pose can be determined using $\mathbf{T}_{i, j}$, we can easily determine the sequence of task instances, $\langle \mathcal{O}_1, \mathcal{O}_2, \dots, \mathcal{O}_k \rangle$.

\noindent
\textbf{Rationale for Deterministic Generation of Task Instances:}
The rationale behind using the deterministic approach for generating new task instances comes from the fact that construction activities are typically assembly tasks where multiple objects with the same geometry are arranged in a pre-determined fashion based on a high-level schematic. For example consider the activity of constructing a wall, the wall is built by arranging multiple bricks of the same geometry as given by the floor-plan. In general, the object geometry and high-level schematics are always available for construction activities and hence, exploiting such structure in these tasks makes sense as it simplifies the problem of generating new task instances for a construction activity.

\noindent
\textbf{Motion Generation:} Once the sequence of task instances have been determined, we can transfer the constant screw constraints that were extracted from the demonstration by following approach proposed in \cite{mahalingam2023humanguided}. This defines the motion of each object in terms of a sequence of $SE(3)$ poses, $\mathcal{GP} = \langle\mathbf{G}_1, \mathbf{G}_2, \dots\rangle$ where each pair of consecutive poses, $\mathbf{G}_{i}$ and $\mathbf{G}_{i+1}$ define a constant screw constraint. We can then use Screw Linear Interpolation (ScLERP) combined with Jacobian pseudo-inverse to compute a motion plan in the joint space \cite{ScLERP}. This ensures that the generated motion plan satisfies the extracted constant screw constraints.
A schematic sketch describing the solution approach is shown in Figure \ref{fig:schematic_sketch}.

\noindent
\textbf{Motion Planning by Exploiting Self-Motion:}
For a $d$-DoF manipulator, at any given configuration $\bm{q}\in\mathbb{R}^{d}$, the spatial Jacobian $\mathbf{J}_s(\mathbf{q})\in\mathbb{R}^{6\times d}$ relates its end-effector spatial velocity $\mathbf{V}_s=\begin{bmatrix}\boldsymbol{v}_s & \boldsymbol{\omega}_s\end{bmatrix}^T\in\mathbb{R}^{6}$ to its joint velocities, $\dot{\bm{q}}\in\mathbb{R}^{d}$
Using the Jacobian pseudoinverse, the joint velocity to track a desired end-effector velocity can be computed as,
\begin{equation}
\dot{\boldsymbol{q}}
\;=\; \mathbf{J}_s^{\dagger}\,\mathbf{V}_s
\text{ where, }
\mathbf{J}_s^{\dagger}=\mathbf{J}_s^{\mathbf{T}}\!\big(\mathbf{J}_s\mathbf{J}_s^{\mathbf{T}})^{-1}.
\label{eq:basic-pinv}
\end{equation}

However, Equation \eqref{eq:basic-pinv} when applied to a redundant manipulator, for example a 7-DoF manipulator, does not control the elbow motion.
Additionally, we can also exploit the self-motion of redundant manipulators which is parametrized by the SEW angle \cite{burdick1989characterization, seraji1989redundant} to avoid hitting joint limits.
In order to control the motion of the elbow, we make use of the SEW Jacobian $\mathbf{J}_{\psi}$ which relates  rate of change of the SEW parameter ($\dot{\psi}$) and manipulator joint velocities, $\dot{\psi} = \mathbf{J}_{\psi}\dot{\boldsymbol{q}}$. By augmenting the task space velocity $\mathbf{V}_s$ with the rate of change of the SEW parameter as shown in \cite{seraji1989redundant}, we can control the motion of the elbow using the augmented Jacobian, $\mathbf{J}_a$.

\begin{equation}
\dot{\boldsymbol{q}} = \boldsymbol{J}_{a}^\dagger
\begin{bmatrix}
    \boldsymbol{v}_s\\
    \boldsymbol{\omega_{s}}\\
    \dot{\psi}
\end{bmatrix},
\boldsymbol{J}_a = \begin{bmatrix}\boldsymbol{J}_s \\ \boldsymbol{J}_{\psi} \end{bmatrix}
\end{equation}

We propose a motion planner with two modes incorporating task space constraints and null space motion as follows:
\begin{align}
\label{eq:sclerp_planner}
\textbf{Mode 1: }&\dot{\boldsymbol{q}}_{\mathrm{ScLERP}} =  \mathbf{J}_{s}^\dagger
\begin{bmatrix}
    \boldsymbol{v}_s\\
    \boldsymbol{\omega_{s}} 
\end{bmatrix} \\
\label{eq:null_space_planner}
\textbf{Mode 2: }&\dot{\boldsymbol{q}}_{\mathrm{SEW}} = \mathbf{J}_{a} ^{\dagger}   \begin{bmatrix}
\mathbf{0}_{6\times 1}\\
    \dot{\psi}
\end{bmatrix}  
\end{align}

In Mode 1, task space constraints are incorporated using ScLERP combined with Jacobian pseudo-inverse method. 
In Mode 2, we move in the null space by introducing non zero $\dot{\psi}$ in the task space velocity and setting the spatial velocity to be zero which causes only the SEW angle to change, thereby ``swinging" the manipulator's elbow. We choose the sign of $\dot{\psi}$ to help us move within the joint position limits. These two modes allow us to plan for motion which satisfies task space constraints while avoiding joint position limits.

The planner works in Mode 1 under normal conditions and switches to Mode 2 only when close to the joint position limits.
We introduce two joint space bounds with margins defined by two parameters $\varepsilon_{\text{in}}, \varepsilon_{\text{out}}$, and the lower/upper joint limit $\ell_i,  u_i$ of each joint $q_{j}$, where
$0<\varepsilon_{\text{out}}<\varepsilon_{\text{in}}<\tfrac12\min_i (u_i-\ell_i)$ which determine the mode of operation of the planner.
\begin{align}
\textbf{Inner bound:}\quad
& \ell_i+\varepsilon_{\text{in}} \;\le\; q_{j} \;\le\; u_i-\varepsilon_{\text{in}}, \label{eq:inner}\\
\textbf{Outer bound:}\quad
& \ell_i+\varepsilon_{\text{out}} \;\le\; q_{j} \;\le\; u_i-\varepsilon_{\text{out}}. \label{eq:outer}
\end{align}

\begin{algorithm}[htbp!]
\caption{ScLERP-SEW Motion Planner}
\label{alg:planner}
\textbf{Input}
$\bm{q}[0], \mathbf{G}_{d}, \varepsilon_{\text{in}}, \varepsilon_{\text{out}}$

\While{$GoalNotReached$}{
    $\mathbf{G}_c = \mathtt{getFK}(\bm{q}[i])$\\
    $\widehat{\bm{\xi}}\theta = \log(\mathbf{G}_d^{-1}\mathbf{G}_c)$\\
    $\dot{\bm{q}} \gets \kappa \mathbf{J}_{s}^\dagger \bm{\xi}\theta$\\
    $\bm{q}[i+1] \gets \bm{q}[i] + \delta_t\dot{\bm{q}}$

    \eIf{$\bm{q}[i+1]$ $ReachedJointLimit$}
    {
    $\psi_d \gets \mathtt{calculateSEWChange} (\bm{q}[i], \varepsilon_{\text{in}}, \varepsilon_{\text{out}})$\\
    \eIf{$\psi_d > 0$}
    {
        $\psi_c = \mathtt{getSEWAngle}(\bm{q}[i])$\\
        \While{$\psi_c \neq \psi_d$}
        {
            $\dot{\bm{q}} \gets \lambda \mathbf{J}_{a}^\dagger 
            \begin{bmatrix}
                \bm{0} \\ \psi_d - \psi_c
            \end{bmatrix}$\\
            $\bm{q}[i+1] \gets \bm{q}[i] + \delta_t\dot{\bm{q}}$\\
            $\psi_c = \mathtt{getSEWAngle}(\bm{q}[i])$\\
            $i = i + 1$
        }
    }
    {
        \textbf{return} $MotionPlanFailed$
    }
    }
    {
    $i = i + 1$
    }
}
\end{algorithm}

Algorithm \ref{alg:planner}, describes the steps involved in determining the sequence of joint configurations $\langle\bm{q}[0], \bm{q}[1], ...\rangle$ that would move the manipulator to the desired task-space goal $\mathbf{G}_{d}\in SE(3)$ while satisfying the constant screw constraints on the end-effector motion to successfully perform the task. The variable $\delta_t \in \mathbb{R}$ represents the time-step.
When no joint limits are hit, the planner uses ScLERP combined with Jacobian pseudoinverse to determine the sequence of joint configurations. When a joint hits its upper bound, it determines how much it can move in the null-space to bring the joints within the inner bound.
The function \texttt{calculateSEWChange()} determines the change in the SEW angle $\psi$ that is required to bring all the joints within the inner bound. If while moving one joint away from the limit, other joints move out of the inner bound and hit the outer bound, then it returns the maximum change in SEW angle $\psi$ that moves the joint position $q_i$ away from the upper bound such that all joints that are close to the upper bound are away from the upper bound by the same distance. If no such motion is possible (such situation arises when two or more joints are at the limits) then it returns $0$. The parameters $\kappa, \lambda \in \mathbb{R}^+$ are time scaling parameters that determine how fast the manipulator moves.

We use this planner to determine the joint space path to go between each consecutive pose in the sequence of guiding poses $\mathcal{GP}$ while satisfying the constant screw constraint.

\section{EXPERIMENTAL RESULTS}

In this section, we provide both the simulation and real-world experimental results for construction tasks of arbitrary length. The simulation experiments were carried out in PyBullet\cite{pybullet}. For all experiments, we use the 7-DoF Franka Emika Panda research robot. We evaluated our approach in simulation across 112 construction activities, and further validated it in real world with 6 construction activities. Here, each activity consisted of either building a wall with multiple bricks or installing multiple ceiling tiles.

First, to answer the question that if a single demonstration is enough for construction tasks with various layouts, we start with a typical construction task, i.e., wall construction, involving multiple pick-and-place sub-tasks. To further evaluate the effectiveness of our method in handling tasks with different kind of constraints, we consider the ceiling-tile installation task, which requires more precise coordination of position and orientation to install the tile.
To show that our approach can be extended to arbitrarily long construction activities, we use a movable manipulator that is mounted on a gantry to construct a long wall consisting of a total of $36$ bricks across $3$ layers in simulation. Please note that we do not plan for the motion of the robot base and are currently just placing it in a neighborhood of the task instance at random to introduce more variation in the task instances.
The key takeaway from this experiment, is that the extracted constraints are invariant to the pose of the robot base or the task-related objects and as long as it is physically feasible for the robot to execute the required motion, the planner can plan for the joint space motion which satisfies the extracted constraints allowing us to perform arbitrarily long construction activities. Moreover, given a high level schematic in the form of a floor plan, we can determine the entire sequence of task instances necessary to perform the entire construction activity showing that our framework can be applied to construction activities of arbitrary length.

For all construction activities, demonstrations were collected in VR. The left half of Figure \ref{fig:simu_experiments} shows the VR environment setup that was used to collect the demonstrations.
In the wall construction activity, the size of brick is: $0.0508 \times 0.0508\times 0.1016$ (meters). In the ceiling-tile installation activity, the tile size is $0.302 \times 0.302\times 0.014$ (meters). Further, we used the following choice of parameters, $\varepsilon_{\text{in}}$ is $0.2$ and $\varepsilon_{\text{out}}$ is $0.01$ for the planner.

To define whether the experiment is success, for each brick-laying trial, the average position error of bricks should be below $0.0075$ m and the orientation error should below $2^{\circ}$ (Yaw angle) with respect to the goal pose. For ceiling tile installation, if the ceiling tile is successfully inserted through the ceiling frame without collision and placed such that it is stably supported on the frame, then it is a success.

For more details about the hardware setup and the conducted experiments, please refer to the following video \texttt{\url{https://youtu.be/SEwtaH--NZg}}.

\begin{figure*}[t]
  \centering
  \vspace{0.6cm}
  \begin{subfigure}[b]{\linewidth}
    \centering
    \includegraphics[width=\linewidth]{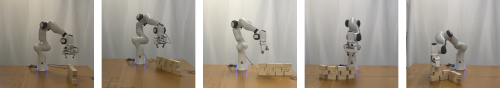}
    \label{fig:subfig1}
  \end{subfigure}

  \begin{subfigure}[b]{\linewidth}
    \centering
    \includegraphics[width=\linewidth]{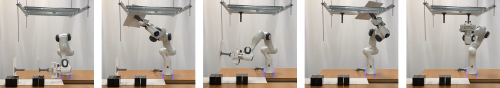}
    \label{fig:subfig2}
  \end{subfigure}
  \caption{\textbf{\textsc{Real World Experiments:}} The first row shows our wall construction experiments with different layouts: three kinds of straight wall, curved wall and corner wall. The second row shows the process of a $1\times2$ ceiling tile installation}
  \label{fig:experiments}
\end{figure*}

\begin{table*}
\centering
\vspace{1.2cm} 
\caption{Comparison between our proposed method and baseline conducted in simulation}
\resizebox{\textwidth}{!}{
\begin{tabular}{c|cc|cc|cc|cc|cc||cc|cc|cc|cc}
\toprule
\multirow{3}{*}{Demonstration}
& \multicolumn{10}{c||}{Wall Construction}
& \multicolumn{8}{c}{Ceiling Tile Installation} \\
\cmidrule(lr){2-11} \cmidrule(lr){12-19}
& \multicolumn{2}{c|}{Layout 1}
& \multicolumn{2}{c|}{Layout 2}
& \multicolumn{2}{c|}{Layout 3}
& \multicolumn{2}{c|}{Layout 4}
& \multicolumn{2}{c||}{Layout 5}
& \multicolumn{2}{c|}{Position 1}
& \multicolumn{2}{c|}{Position 2}
& \multicolumn{2}{c|}{Position 3}
& \multicolumn{2}{c}{Position 4} \\
\cmidrule(lr){2-19}
& Ours & Baseline
& Ours & Baseline
& Ours & Baseline
& Ours & Baseline
& Ours & Baseline
& Ours & Baseline
& Ours & Baseline
& Ours & Baseline
& Ours & Baseline \\
\midrule
Demo 1 &  15&  0&  15&  15&  9&  9&  12&  12&  18&  15& \cmark & \xmark & \cmark & \xmark & \cmark & \xmark & \cmark & \xmark \\
Demo 2 &  15&  15&  15&  12&  15&  15&  12&  12&  21&  18& \cmark & \xmark & \cmark & \xmark & \cmark & \xmark & \cmark & \xmark \\
Demo 3 &  15&  12&  12&  12&  15&  9&  12&  11&  18&  18& \cmark & \xmark & \xmark & \xmark & \cmark & \xmark & \cmark & \xmark \\
Demo 4 &  15&  15&  15&  12&  21&  18&  12&  12&  21&  18& \cmark & \xmark & \cmark & \xmark & \cmark & \xmark & \cmark & \xmark \\
\bottomrule
\end{tabular}
}

\label{table:baseline compare}
\end{table*}
\subsection{Simulation Experiments}

\noindent
\textbf{Fixed-Base Experiments:}
In this part, we evaluate our methods across two construction activities which contain repetitive sub-tasks: (\romannumeral 1) brick wall construction. (\romannumeral 2) ceiling tile installation.

For task (\romannumeral 1) we consider building a wall with $5$ different layouts by arranging bricks one by one: (1) A straight wall parallel to the $\bm x$-axis, (2) A wall identical to the previous experiment but rotated $30^{\circ}$ with respect to the $\bm x$-axis, (3) A straight wall parallel to the $\bm y$-axis, (4) A $120^{\circ}$ circular wall, (5) A corner wall. All walls consist of $3$ layers of bricks with a total of 12 bricks. For task (\romannumeral 2) 2 ceiling tiles were installed in a $1\times2$ ceiling frame.

\noindent
\textbf{Moving-Base Experiments:}
In this part, by assuming the manipulator can move freely in $\bm{xy}$-plane, we increase the scale of the experiments to two longer activities: (\romannumeral 1) a $3$-layer long wall aligned with $\bm y$-axis, consisting of $36$ bricks, (\romannumeral 2) a $2\times2$ ceiling tile installation task.

For the long wall construction experiment, the manipulator moves along $y$-axis after placing $3$ bricks. For the ceiling tile installation task, the manipulator moves after completing each tile installation.
In both the arbitrary long construction tasks, each manipulator base movement changes the relative pose between the manipulator and the new task instances, this setting allows us to evaluate whether the motion planner can robustly handle variations in relative pose. 

The right half of Figure \ref{fig:simu_experiments} shows few of the experiments conducted in simulation.


\noindent
\textbf{Baseline Comparison:}
To access the ability of the proposed SEW-ScLERP based motion planner, we conducted a comparison with the approach proposed in \cite{mahalingam2023humanguided} as the baseline. The comparison was conducted in simulation with the fixed-base setting. The results of the conducted experiments for the wall building with the $5$ layouts and ceiling tile installation are summarized in Table \ref{table:baseline compare}. For the wall construction task, the table lists the number of bricks that were successfully placed before the planner hit the joint limits. For the ceiling tile task, it shows if the execution succeeded or failed. We can clearly see that the proposed approach performs better than the baseline across all construction activities. For the wall construction activity, the proposed approach achieves a higher cumulative number for successful bricks placed ($303$ vs. $260$). For the ceiling tile installation task, the proposed approach succeeded in $15$ of the $16$ trials whereas the baseline failed across all the trials.

\subsection{Real-World Experiments}

To further validate our proposed approach, we conducted real world experiments in a lab environment using a fixed base manipulator. We performed the wall construction activity with the $5$ different layouts and the ceiling tile installation activity. Figure \ref{fig:experiments} shows the different real world experiments that were conducted. For the wall construction all the layouts consisted of $3$ layers with a total of $12$ bricks each. For the ceiling tile installation, we had a $2\times1$ ceiling tile framing setup right above the robot and the robot inserts two tiles one by one into each frame. All of the conducted trials were successful showing that the proposed framework of using VR to acquire manipulation skills for performing construction activities is robust when dealing with long activities which consist of complex motion constraints.

\section{Conclusion and Future Work}
In this paper, we presented a novel framework for automating arbitrarily long construction activities by generalizing from a single step-level demonstration collected in a VR environment. Our approach segments the recorded object trajectory in $SE(3)$ into constant screw motions, extracts the underlying kinematic constraints, and reuses these parameterized motion primitives to generate motion plans for new task instances. We demonstrated the effectiveness of this framework on a few representative construction activities, showing that it can scale from a single elemental demonstration to the completion of arbitrarily long, layout-dependent construction activities. This contribution represents a significant step toward data-efficient automation of repetitive, constraint-rich tasks in construction robotics. 

\bibliographystyle{IEEEtran} 
\bibliography{references}

\end{document}